# Mini-Bucket Heuristics for Improved Search

Kalev Kask and Rina Dechter

Department of Information and Computer Science
University of California, Irvine, CA 92697-3425
{kkask,dechter}@ics.uci.edu

## Abstract

The paper is a second in a series of two papers evaluating the power of a new scheme that generates search heuristics mechanically. The heuristics are extracted from an approximation scheme called mini-bucket elimination that was recently introduced. The first paper introduced the idea and evaluated it within Branch-and-Bound search. In the current paper the idea is further extended and evaluated within Best-First search. The resulting algorithms are compared on coding and medical diagnosis problems, using varying strength of the mini-bucket heuristics.

Our results demonstrate an effective search scheme that permits controlled tradeoff between preprocessing (for heuristic generation) and search. Best-first search is shown to outperform Branch-and-Bound, when supplied with good heuristics, and sufficient memory space.

## 1 Introduction

The paper is a second in a series of two papers evaluating the power of a new scheme that generates search heuristics mechanically. In the first paper [Kask and Dechter, 1999a], we proposed a new scheme that uses the mini-bucket approximation methods to generate heuristics for search algorithms. Since the mini-bucket's approximation accuracy is controlled by a bounding parameter, it allows heuristics having varying degrees of accuracy and results in a spectrum of search algorithms that can tradeoff heuristic computation and search.

The idea was studied using a branch and bound search for finding the most probable explanation (MPE) in Bayesian networks. Empirical evaluations demonstrated good performance, superior to algorithms such as bucket elimination or join-tree clustering, while improving on mini-bucket approximations.

In the current paper we explore the power of the mini-bucket heuristics within *Best-First* search. Since, as shown, these heuristics are admissible and monotonic, their use within Best-First search yields A* type algorithms whose properties are well understood; the algorithm is guaranteed to terminate with an optimal solution; when provided with more powerful heuristics it explores a smaller search space, but otherwise it requires substantial space. It is also known that Best-First algorithms are optimal. Namely, when given the same heuristic information, Best-First search is the most efficient algorithm in terms of the size of the search space it explores [Dechter and Pearl, 1985]. In particular, Branch-and-Bound will expand any node that is expanded by Best-First (up to some tie breaking conditions) and, in many cases it explores a larger space. Still, Best-First may occasionally fail because of its memory requirements.

The question we investigate here is to what extent the mini-bucket heuristics can facilitate the solution of larger and harder problems by Best-First search, and how Best-First is compared with Branch-and-Bound, when both have access to the same heuristic information.

Mini-bucket is a class of parameterized approximation algorithms based on the bucket-elimination framework [Dechter, 1996]. The approximation uses a controlling parameter which allows adjustable levels of accuracy and efficiency [Dechter and Rish, 1997]. The algorithms were presented and analyzed for deterministic and probabilistic tasks such as finding the most probable explanation ($MPE$), belief updating and finding the maximum a posteriori hypothesis. Encouraging empirical results were reported on randomly generated noisy-or networks, on medical-diagnosis CPCS networks, and on coding problems [Rish et al., 1998]. In some cases however the approximation is seriously suboptimal even when using the highest feasible accu-racy level. This can be determined by an error bound produced by the mini-bucket scheme.

Branch-and-Bound searches the space of partial assignments in a depth-first manner. It will expand a partial assignment only if its upper-bounding heuristic function is larger than the currently known lower bound solution. The virtue of branch-and-bound is that it requires a limited amount of memory and can be used as an anytime scheme; whenever interrupted, branch-and-bound outputs the best solution found so far. Best-First explores the search space in uniform frontiers of partial instantiations, each having the same value for the evaluation functions, while progressing in waves of decreasing values.

In this paper, a Best-First algorithm with Mini-Bucket heuristics (BFMB) is evaluated empirically and compared with a Branch and Bound algorithm using Mini-Bucket heuristics (BBMB), with mini-bucket approximation scheme and with iterative belief propagation, over test problems such as coding networks, noisy-or networks and CPCS networks.

We show that Best-First frequently outperforms Branch and Bound, whenever Best-First terminates. Namely, when the heuristics were strong enough, if given enough time and when memory problems were not encountered. Both search methods exploit heuristics strength in a similar manner; on all problem classes, the optimal tradeoff point between heuristic generation and search lies in an intermediate range of the heuristics' strength. This optimal point gradually increases towards stronger heuristics, as problems become larger and harder.

Section 2 provides preliminaries and background on the mini-bucket algorithms. Section 3 describes the heuristic function which is built on top of the mini-bucket algorithm, proves its properties and embed the heuristic within Best-First search. Section 4 presents empirical evaluations while section 5 provides discussion and conclusions.

### 1.1 Related work

Our approach applies the paradigm that heuristics can be generated by consulting relaxed models, suggested in [Pearl, 1984]. The mini-bucket heuristics can also be viewed as an extension of bounded constraint propagation algorithms that were investigated in the constraint community in the last decade [Dechter, 1992].

Here is some related work for finding the most probable explanation in Bayesian networks. It is known that solving the MPE task is NP-hard. Complete algorithms for MPE use either the *cycle cutset* (also called *conditioning*) technique or the *join-tree-clustering* technique [Pearl, 1988] or bucket-elimination scheme [Dechter, 1996]. However, these methods work well only if the network is sparse enough to allow small cutsets or small clusters. Following Pearl's stochastic simulation algorithms for the MPE task [Pearl, 1988], the suitability of Stochastic Local Search (SLS) algorithms for MPE was studied in the context of medical diagnosis applications [Peng and Reggia, 1989] and more recently in [Kask and Dechter, 1999b]. Best first search algorithms were also proposed in [Shimony and Charniak, 1991] as well as algorithms based on linear programming [Santos, 1991].

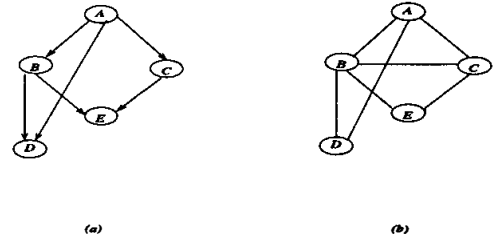

Figure 1: a) A belief network $P(e,d,c,b,a) = P(e|c,b)P(d|b,a)P(b|a)P(c|a)P(a)$, b) Its moral graph

## 2 Background

### 2.1 Notation and definitions

*Belief Networks* provide a formalism for reasoning about partial beliefs under conditions of uncertainty. They are defined by a directed acyclic graph over nodes representing random variables of interest.

DEFINITION 2.1 (**Belief Networks**) *Given a set, $X = \{X_1, \ldots, X_n\}$ of random variables over multivalued domains $D_1, \ldots, D_n$, a belief network is a pair $(G, P)$ where $G$ is a directed acyclic graph and $P = \{P_i\}$. $P_i = \{P(X_i|pa(X_i))\}$ are conditional probability matrices associated with $X_i$. The set $pa(X_i)$ is called the parent set of $X_i$. An assignment $(X_1 = x_1, \ldots, X_n = x_n)$ can be abbreviated to $x = (x_1, \ldots, x_n)$. The BN represents a probability distribution $P(x_1, \ldots, x_n) = \Pi_{i=1}^n P(x_i|x_{pa(X_i)})$, where, $x_S$ is the projection of $x$ over a subset $S$. An evidence set $e$ is an instantiated subset of variables. The argument set of a function $h$ are denoted $S(h)$. An example of a belief network is given in Figure 1.*

DEFINITION 2.2 (**Most Probable Explanation**) *Given a belief network and evidence $e$, the Most Probable Explanation (MPE) task is to find an assignment $(x_1^o, \ldots, x_n^o)$ such that*

$$P(x_1^o, \ldots, x_n^o) = max_{X_1, \ldots, X_n} \prod_{k=1}^{n} P(X_k \mid pa(X_k), e)$$

DEFINITION 2.3 (graph concepts) *An ordered graph is a pair $(G, d)$ where $G$ is an undirected graph and $d = X_1, \ldots, X_n$ is an ordering of the nodes. The width of a node in an ordered graph is the number of its earlier neighbors. The width of an ordering $d$, $w(d)$, is the maximum width over all nodes. The induced width of an ordered graph, $w^*(d)$, is the width of the induced ordered graph obtained by processing the nodes recursively, from last to first; when node $X$ is processed, all its earlier neighbors are connected. The moral graph of a directed graph $G$ is the undirected graph obtained by connecting the parents of all the nodes in $G$ and then removing the arrows.*

## 2.2 Bucket and mini-bucket algorithms

*Bucket elimination* is a unifying algorithmic framework for dynamic-programming algorithms applicable to probabilistic and deterministic reasoning [Bertele and Brioschi, 1972, Dechter, 1996]. The input to a bucket-elimination algorithm consists of a collection of functions or relations (e.g., clauses for propositional satisfiability, constraints, or conditional probability matrices for belief networks). Given a variable ordering, the algorithm partitions the functions into buckets, each associated with a single variable. A function is placed in the bucket of its latest argument in the ordering. The algorithm has two phases. During the first, top-down phase, it processes each bucket, from the last variable to the first. Each bucket is processed by a variable elimination procedure that computes a new function which is placed in a lower bucket. For MPE, the bucket procedure generates the product of all probability matrices and maximizes over the bucket's variable. During the second, bottom-up phase, the algorithm constructs a solution by assigning a value to each variable along the ordering, consulting the functions created during the top-down phase.

THEOREM 2.1 *[Dechter, 1996] The time and space complexity of the algorithm Elim-MPE, the bucket elimination algorithm for MPE, are exponential in the induced width $w^*(d)$ of the network's ordered moral graph along the ordering $d$.* □

*Mini-bucket elimination* is an approximation designed to avoid the space and time problem of full bucket elimination. In each bucket, all the functions are partitioned into smaller subsets called mini-buckets which are processed independently. Here is the rationale. Let $h_1, \ldots, h_j$ be the functions in $bucket_p$.

---

Algorithm **Approx-MPE(i)** (MB(i))
Input: A belief network $BN = \{P_1, \ldots, P_n\}$; ordering $d$;
Output: An upper bound on the MPE, an assignment and the set of ordered augmented buckets.
1. Initialize: Partition matrices into buckets. Let $S_1, \ldots, S_j$ be the subset of variables in $bucket_p$ on which matrices (old or new) are defined.
2. (**Backward**) For $p \leftarrow n$ downto 1, do
• If $bucket_p$ contains $X_p = x_p$, assign $X_p = x_p$ to each $h_i$ and put each in appropriate bucket.
• else, for $h_1, h_2, \ldots, h_j$ in $bucket_p$, generate an $(i)$-partitioning, $Q' = \{Q_1, \ldots, Q_r\}$. For each $Q_l \in Q'$ containing $h_{l_1}, \ldots h_{l_t}$ generate function $h^l$, $h^l = max_{X_p} \Pi_{i=1}^{t} h_{l_i}$. Add $h^l$ to the bucket of the largest-index variable in $U_l \leftarrow \bigcup_{i=1}^{p} S(h_{l_i}) - \{X_p\}$.
3. (**Forward**) For $i = 1$ to $n$ do, given $x_1, \ldots, x_{p-1}$ choose a value $x_p$ of $X_p$ that maximizes the product of all the functions in $X_p$'s bucket.
4. Output the ordered set of augmented buckets, an upper bound and a lower bound assignment.

Figure 2: Algorithm *Approx-MPE(i)*

When *Elim-MPE* processes $bucket_p$, it computes the function $h^p$: $h^p = max_{X_p} \Pi_{i=1}^{j} h_i$. The mini-bucket algorithm, on the other hand, creates a partitioning $Q' = \{Q_1, \ldots, Q_r\}$ where the mini-bucket $Q_l$ contains the functions $h_{l_1}, \ldots, h_{l_k}$. The approximation will compute $g^p = \Pi_{l=1}^{r} max_{X_p} \Pi_{i} h_{l_i}$. Clearly, $h^p \leq g^p$. Thus, the algorithm computes an upper bound on the probability of the MPE assignment. Subsequently, the algorithm computes an assignment that provides a lower bound. The quality of the upper bound depends on the degree of the partitioning into mini-buckets. Given a bound parameter $i$, the algorithm creates an $i$-partitioning, where each mini-bucket includes no more than $i$ variables. Algorithm *Approx-MPE(i)* (sometimes called MB(i)), described in Figure 2, is parameterized by this $i$-bound. The algorithm outputs not only an upper bound on the $MPE$ and an assignment (whose probability yields a lower bound), but also the collection of augmented buckets. By comparing the upper bound to the lower bound we can always have a bound on the error for the given instance.

The algorithm's complexity is time and space $O(exp(i))$ where $i \leq n$. When the bound $i$ is large enough (i.e. when $i \geq w^*$), the mini-bucket algorithm coincides with the full bucket elimination algorithm Elim-MPE. In summary,

THEOREM 2.2 *[Dechter and Rish, 1997] Algorithm Approx-MPE(i) generates an upper bound on the exact MPE and its time and space complexity is exponential in its bound $i$.*

**Example 2.3** *Figure 3(b) illustrates how algorithms Elim-MPE and* Approx-MPE(i) *for $i = 3$ process*

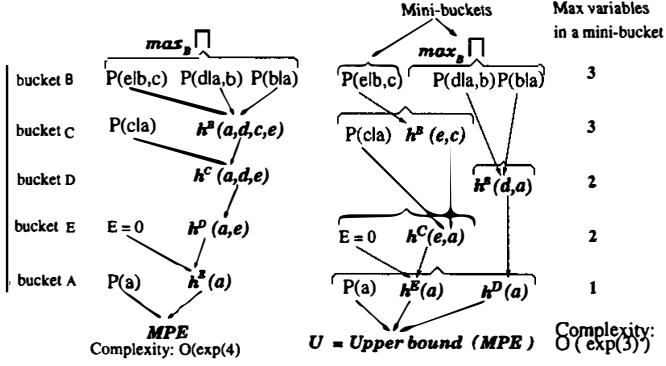

(a) A trace of *Elim-MPE*   (b) A trace of *Approx-MPE(3)*

Figure 3: Execution of *Elim-MPE* and *Approx-MPE*

the network in Figure 1(a) along the ordering $(A, E, D, C, B)$. Algorithm Elim-MPE records new functions $h^B(a,d,c,e)$, $h^C(a,d,e)$, $h^D(a,e)$, and $h^E(a)$. Then, in the bucket of $A$, it computes $MPE = \max_a P(a)h^E(a)$. Subsequently, an MPE assignment $(A = a', B = b', C = c', D = d', E = 0)$ ($E = 0$ is an evidence) is computed for each variable from $A$ to $B$ by selecting a value that maximizes the product of functions in the corresponding bucket, conditioned on the previously assigned values. Namely, $a' = \arg\max_a P(a)h^E(a)$, $e' = 0$, $d' = \arg\max_d h^C(a',d,e=0)$, and so on. The approximation Approx-MPE(3) splits bucket $B$ into two mini-buckets each containing no more than 3 variables, and generates $h^B(e,c)$ and $h^B(d,a)$. An upper bound on the MPE value is computed by $\max_a P(a) \cdot h^E(a) \cdot h^D(a)$. A suboptimal MPE tuple is computed similarly to MPE tuple by assigning a value to each variable that maximizes the product of functions in the corresponding bucket, given the assignments to the previous variables.

## 3 Heuristic Search with Mini-Bucket

### 3.1 Notation

In the following discussion we will assume that the mini-bucket algorithm was applied to a belief network using a given variable ordering $d = X_1, ..., X_n$, and that the algorithm outputs an ordered set of augmented buckets bucket 1,...,bucket $p$,...,bucket $n$, containing the input functions and the newly generated functions. Relative to such an ordered set of augmented buckets we use the following convention.

- $P_{p_j}$ denotes the input conditional probability matrices placed in bucket $p$, (namely, its highest-ordered variable is $X_p$).

- $h_{p_j}$ denotes an arbitrary function in bucket $p$ generated by the mini-bucket algorithm.

- $h_j^p$ denotes a function created by the $j$-th mini-bucket in bucket $p$.

- $\lambda_{p_j}$ denotes an arbitrary function in bucket $p$.

We denote by $buckets(1..p)$ the union of all functions in the bucket of $X_1$ through the bucket of $X_p$. Remember that $S(f)$ denotes the set of arguments of function $f$.

### 3.2 The Heuristic Function

The idea, first presented in [Kask and Dechter, 1999a], is given here for the completeness of the presentation. We will show that the new functions recorded by the mini-bucket algorithm can be used to express upper bounds on the most probable extension of any partial assignment. Therefore, they can serve as heuristics in an evaluation function which guides a *Best-First* search or as an upper bounding function for pruning *Branch-and-Bound* search.

**DEFINITION 3.1 (Exact Evaluation Function)**
Let $\bar{x} = \bar{x}^p = (x_1, ..., x_p)$. The probability of the most probable extension of $\bar{x}^p$, denoted $f^*(\bar{x}^p)$ is:

$$max_{\{X_{p+1},...,X_n | X_i = x_i, \forall i, 1 \leq i \leq p\}} \prod_{k=1}^{n} P(X_k | pa(X_k), e)$$

The above product defining $f^*$ can be divided into two smaller products expressed by the functions in the ordered augmented buckets. In the first product all the arguments are instantiated, and therefore the maximization operation is applied to the second product only. Denoting $g(\bar{x}) = \prod_{P_i \in buckets(1..p)} P_i(\bar{x}_{S(P_i)})$ and $H^*(\bar{x}) = max_{\{X_{p+1},...,X_n | X_i = x_i, \forall 1 \leq i \leq p\}} \prod_{P_i \in buckets(p+1..n)} P_i$, we get $f^*(\bar{x}) = g(\bar{x}) \cdot H^*(\bar{x})$. During search, the $g$ function can be evaluated over the partial assignment $\bar{x}^p$, while $H^*$ can be estimated by a heuristic function $H$ defined next.

**DEFINITION 3.2** *Given an ordered set of augmented buckets, the heuristic function $H(\bar{x}^p)$, is the product of all the $h$ functions that satisfy the following two properties: 1) They are generated in buckets $(p+1, ..., n)$, and 2) They reside in buckets 1 through $p$. Namely, $H(\bar{x}^p) = \prod_{i=1}^{p} \prod_{h_j^k \in bucket_i} h_j^k$, where $k > p$, (i.e. $h_j^k$ is generated by a bucket processed before bucket $p$.)*

The following proposition shows how $g(\bar{x}^{p+1})$ and $H(\bar{x}^{p+1})$ can be updated recursively based on $g(\bar{x}^p)$ and $H(\bar{x}^p)$ and functions residing in bucket $p+1$.

**Proposition 1** *Given a partial assignment $\bar{x}^p = (x_1, \ldots x_p)$, both $g(\bar{x}^p)$ and $H(\bar{x}^p)$ can be computed recursively by*

$$g(\bar{x}^p) = g(\bar{x}^{p-1}) \cdot \Pi_j P_{p_j}(\bar{x}^p_{S(P_{p_j})}) \qquad (1)$$

$$H(\bar{x}^p) = H(\bar{x}^{p-1}) \cdot \Pi_k h_{p_k} / \Pi_j h_j^p \qquad (2)$$

THEOREM 3.1 (**Mini-Bucket Heuristic**) *For every partial assignment $\bar{x} = \bar{x}^p = (x_1, \ldots, x_p)$, of the first $p$ variables, the evaluation function $f(\bar{x}^p) = g(\bar{x}^p) \cdot H(\bar{x}^p)$ is: 1) Admissible - it never underestimates the probability of the best extension of $\bar{x}^p$. 2) Monotonic, namely $f(\bar{x}^{p+1})/f(\bar{x}^p) \leq 1$.*

**Proof.** To prove monotonicity we will use the recursive equations (1) and (2) from Proposition 1. For any $\bar{x}^p$ and any value $v$ in the domain of $X_{p+1}$, we have

$$f(\bar{x}^p, v)/f(\bar{x}^p) = (g(\bar{x}^p, v) \cdot H(\bar{x}^p, v))/(g(\bar{x}^p) \cdot H(\bar{x}^p)) =$$

$$= \Pi_i \lambda_{(p+1)_i}(\bar{x}^p, v)/\Pi_j h_j^{p+1}(\bar{x}^p).$$

Since $h_j^{p+1}(\bar{x}^p)$ is computed for each mini-bucket $j$ in bucket $(p+1)$ by maximizing over variable $X_{p+1}$, (eliminating variable $X_{p+1}$), we get

$$\Pi_i \lambda_{(p+1)_i}(\bar{x}^p, v) \leq \Pi_j h_j^{p+1}(\bar{x}^p, v)).$$

Thus, $f(\bar{x}^p, v) \leq f(\bar{x}^p)$, concluding the proof of monotonicity.

The proof of admissibility follows from monotonicity. It is well known that if a heuristic function is monotone and if it is exact for a full solution (which is our case, since the heuristic is the constant 1 on a full solution) then it is also admissible [Pearl, 1984]. □

### 3.3 Search with Mini-Bucket Heuristics

The tightness of the upper bound generated by mini-bucket approximation depends on its $i$-bound parameter. Larger values of $i$ generally yield better upper-bounds, but require more computation. Therefore, both Branch-and-Bound search and Best-First search, if parameterized by $i$, allow a controllable tradeoff between preprocessing and search, or between heuristic strength and its overhead.

In Figures 4 and 5 we present algorithms BBMB and BFMB. Both algorithms are initialized by running the mini-bucket approximation algorithm that produces a set of ordered augmented buckets.

Branch and bound with mini-bucket heuristics (BBMB) traverses the search space in a depth-first manner, instantiating variables from first to last.

---

**Algorithm BBMB(i)**
**Input:** A belief network $BN = \{P_1, ..., P_n\}$; ordering $d$; time bound $t$.
**Output:** An MPE assignment, or a lower bound and an upper-bound on the MPE.
1. **Initialize:** Run MB(i) algorithm which generates a set of ordered augmented buckets and an upper-bound on MPE. Set lower bound $L$ to 0. Set current variable index $p$ to 0.
2. **Search:** Execute the following procedure until variable $X_1$ has no legal values left, or out of time, in which case output the current best solution.
• **Expand:** Given a partial instantiation $\bar{x}^p$, compute all partial assignments $\bar{x}^{p+1} = (\bar{x}^p, v)$ for each value $v$ of $X_{p+1}$. For each node $\bar{x}^{p+1}$ compute its heuristic value $f(\bar{x}^{p+1}) = g(\bar{x}^{p+1}) \cdot H(\bar{x}^{p+1})$ using
$g(\bar{x}^{p+1}) = g(\bar{x}^p) \cdot \Pi_j P_{p+1_j}$ and
$H(\bar{x}^{p+1}) = H(\bar{x}^p) \cdot \Pi_k h_{p+1_k} / \Pi_j h_j^{p+1}$.
Prune those assignments $\bar{x}^{p+1}$ for which $f(\bar{x}^{p+1})$ is smaller than the lower bound $L$.
• **Forward:** If $X_{p+1}$ has no legal values left, goto Backtrack. Otherwise let $\bar{x}^{p+1} = (\bar{x}^p, v)$ be the best extension to $\bar{x}^p$ according to $f$. If $p + 1 = n$, then set $L = f(\bar{x}^{p+1})$ and goto Backtrack. Otherwise remove $v$ from the list of legal values. Set $p = p + 1$ and goto Expand.
• **Backtrack:** If p = 1, Exit. Otherwise set $p = p - 1$ and repeat the Forward step.

Figure 4: Algorithm *BBMB(i)*

---

Throughout the search, the algorithm maintains a lower bound on the probability of the MPE assignment, which corresponds to the probability of the best full variable instantiation found thus far. When the algorithm processes variable $X_p$, all the variables preceding $X_p$ in the ordering are already instantiated, so it can compute the heuristic value $f(\bar{x}^{p-1}, X_p = v) = g(\bar{x}^{p-1}, v) \cdot H(\bar{x}^p, v)$ for each extension $X_p = v$. The algorithm prunes all values $v$ whose heuristic estimate (upper bound) $f(\bar{x}^p, X_p = v)$ is less or equal to the current best lower bound, because such a partial assignment $(x_1, \ldots x_{p-1}, v)$ cannot be extended to an improved full assignment. The algorithm assigns the best value $v$ to variable $X_p$, and proceeds to variable $X_{p+1}$, and when variable $X_p$ has no values left, it backtracks to variable $X_{p-1}$. Search terminates when it reaches a time-bound or when the first variable has no values left. In the latter case, the algorithm has found an optimal solution.

Algorithm Best-First with mini-bucket heuristics (BFMB), starts by adding a dummy node $x_0$ to the list of open nodes. Each node corresponds to a partial assignment $\bar{x}^p$ and has an associated heuristic value $f(\bar{x}^p)$. Initially $f(x_0) = 1$. The basic step of the algorithm consists of selecting an assignment $\bar{x}^p$ from the list of open nodes having the highest heuristic value $f(\bar{x}^p)$, expanding it by computing all partial assignments $(\bar{x}^p, v)$ for all values $v$ of $X_{p+1}$, and adding them

```
Algorithm BFMB(i)
Input: A belief network BN = {P_1, ..., P_n}; ordering d;
time bound t.
Output: An MPE assignment or just an upper bound
and a lower bound (produced by mini-bucket).
1. Initialize: Run MB(i) algorithm which generates a set
of augmented buckets, an upper-bound and a lower bound
assignment. Insert a dummy node x̄_0 in the set L of open
nodes. Set f(x̄_0) to 1.
2. Search:
• If out of time, output mini-bucket assignment.
• Select and remove a node x̄^p with the largest heuristic
value f(x̄^p) from the set of open nodes L.
• If n = p then x̄^p is an optimal solution. Exit.
• Expand x̄^p by computing all child nodes (x̄^p, v) for each
value v in the domain of X_{p+1}. For each node x̄^{p+1} com-
pute its heuristic value f(x̄^{p+1}) = g(x̄^{p+1})H(x̄^{p+1}), where
g(x̄^p) = g(x̄^{p-1}) · Π_j P_{p_j} and
H(x̄^p) = H(x̄^{p-1}) · Π_k h_{p_k}/Π_j h_j^p

• Add all nodes (x̄^p, v) to L and goto Search.
```

Figure 5: Algorithm *BFMB(i)*

to the list of open nodes. The algorithm terminates when it selects a complete assignment for expansion which is guaranteed to be optimal.

## 4 Experimental Methodology

We tested the performance of our scheme on three types of networks - random coding networks, Noisy-OR networks and CPCS networks. On each problem we ran both BBMB(i) and BFMB(i) using mini-bucket approximation with various i-bounds. On random coding networks we also ran for comparison the Iterative Belief Propagation (IBP) [Pearl, 1988], the best algorithm known for probabilistic decoding.

We used the min-degree heuristic for computing the ordering of variables. It places a variable with the smallest degree at the end of the ordering, connects all of its neighbors, removes the variable from the graph and repeats the whole procedure.

We treat all algorithms as approximation algorithms. Algorithms BBMB and BFMB, if allowed to run until completion will solve all problems exactly. However, since we use a time-bound, both algorithms may return suboptimal solutions, especially for harder and larger instances. BBMB outputs its best solution while BFMB, if interrupted, outputs the mini-bucket solution. Consequently BFMB is effective only as a complete algorithm.

The main measure of performance we used is the accuracy ratio $opt = P_{alg}/P_{MPE}$ between the probability of the solution found by the test algorithm ($P_{alg}$) and the probability of the optimal solution ($P_{MPE}$), given a fixed time bound, whenever $P_{MPE}$ is available. We also record the running time of each algorithm.

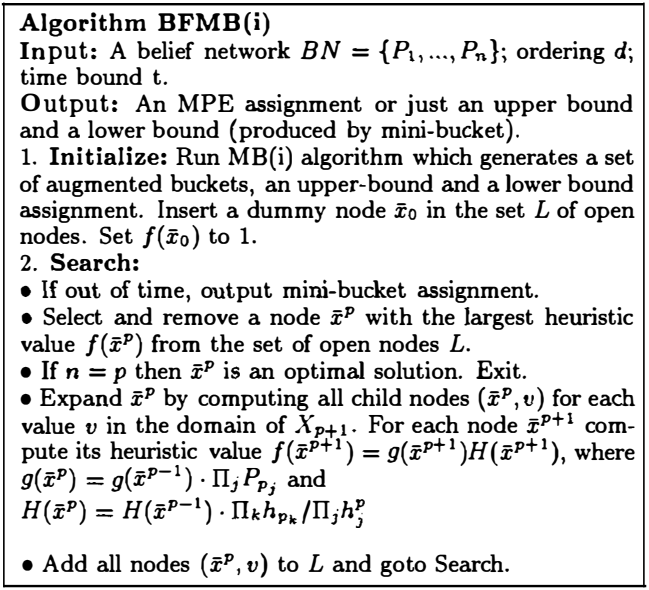

Figure 6: Belief network for structured (10,5) block code with parent set size P=3

We recorded the distribution of problems with respect to accuracy *opt* over 5 predefined ranges : $opt \geq 0.95$, $opt \geq 0.5$, $opt \geq 0.2$, $opt \geq 0.01$ and $opt < 0.01$. However, because of space restrictions, we report only the number of problems that fall in the accuracy range $opt \geq 0.95$. Problems in this range were solved optimally.

In addition, during the execution of both BBMB and BFMB we also stored the current lower bound $L$ at regular time intervals. This allows reporting of accuracy as a function of time.

### 4.1 Random Coding Networks

Our random coding networks fall within the class of *linear block codes*. They can be represented as four-layer belief networks (Figure 6). The second and third layers correspond to input information bits and parity check bits respectively. Each parity check bit represents an XOR function of input bits $u_i$. Input and parity check nodes are binary while the output nodes are real-valued. In our experiments each layer has the same number of nodes because we use code rate of R=K/N=1/2, where K is the number of input bits and N is the number of transmitted bits.

Given a number of input bits K, number of parents P for each XOR bit and channel noise variance $\sigma^2$, a coding network structure is generated by randomly picking parents for each XOR node. Then we simulate an input signal by assuming a uniform random distribution of information bits, compute the corresponding values of the parity check bits, and generate an assignment to the output nodes by adding Gaussian noise to each information and parity check bit. The decoding algorithm takes as input the coding network and the observed real-valued output assignment and recovers the original input bitvector by computing or approximating an MPE assignment. In our experiments all coding networks were generated by randomly picking

| N=100 K=50 σ | opt | MB BBMB BFMB i=2 # / time | MB BBMB BFMB i=6 # / time | MB BBMB BFMB i=10 # / time | MB BBMB BFMB i=14 # / time | IBP |
|---|---|---|---|---|---|---|
| 0.22 | >0.95 | 86/0.04<br>100/0.06<br>100/0.06 | 89/0.06<br>100/0.14<br>100/0.08 | 97/0.32<br>100/0.33<br>100/0.34 | 99/3.26<br>100/3.26<br>100/3.27 | 100/<br>0.09 |
| 0.28 | >0.95 | 74/0.04<br>99/0.38<br>100/0.13 | 70/0.06<br>100/0.40<br>100/0.10 | 86/0.34<br>100/0.40<br>100/0.37 | 97/3.13<br>100/3.14<br>100/3.39 | 100/<br>0.09 |
| 0.32 | >0.95 | 45/0.05<br>96/0.94<br>100/0.13 | 56/0.06<br>100/0.78<br>100/0.10 | 71/0.34<br>100/0.40<br>100/0.37 | 81/3.34<br>100/3.39<br>100/3.39 | 99/<br>0.09 |
| 0.40 | >0.95 | 14/0.04<br>95/3.13<br>99/0.87 | 20/0.06<br>99/2.20<br>100/0.64 | 44/0.32<br>100/0.70<br>100/0.48 | 62/3.07<br>100/3.11<br>100/3.10 | 90/<br>0.07 |
| 0.51 | >0.95 | 3/0.04<br>77/12.0<br>71/9.05 | 8/0.06<br>92/8.15<br>88/6.84 | 13/0.34<br>100/2.52<br>99/2.78 | 18/3.38<br>100/4.00<br>100/4.07 | 32/<br>0.08 |

Table 1: Random coding, N=100 K=50. 100 samples.

| N=100 K=50 σ | MB BBMB BFMB i=2 | MB BBMB BFMB i=6 | MB BBMB BFMB i=10 | MB BBMB BFMB i=14 | IBP |
|---|---|---|---|---|---|
| 0.22 | 0.006000<br>0.000200<br>0.000200 | 0.004600<br>0.000200<br>0.000200 | 0.001000<br>0.000200<br>0.000200 | 0.000400<br>0.000200<br>0.000200 | 0.000200 |
| 0.28 | 0.018200<br>0.001400<br>0.000200 | 0.021200<br>0.000200<br>0.000200 | 0.004800<br>0.000200<br>0.000200 | 0.000100<br>0.000200<br>0.000200 | 0.000200 |
| 0.32 | 0.044800<br>0.007200<br>0.002200 | 0.036200<br>0.002200<br>0.002200 | 0.025600<br>0.002200<br>0.002200 | 0.014800<br>0.002200<br>0.002200 | 0.002200 |
| 0.40 | 0.099600<br>0.019400<br>0.011600 | 0.099600<br>0.012000<br>0.008800 | 0.062800<br>0.008800<br>0.008800 | 0.040600<br>0.008800<br>0.008800 | 0.008800 |
| 0.51 | 0.191600<br>0.098000<br>0.097400 | 0.185200<br>0.083000<br>0.082200 | 0.163000<br>0.076200<br>0.076600 | 0.148600<br>0.076200<br>0.076200 | 0.080000 |

Table 2: Random coding BER, N=100 K=50. 100 samples.

4 parents for each XOR bit.

Tables 1 through 4 report on random coding networks. In addition to BBMB and BFMB, we also ran Iterative Belief Propagation (IBP) [Pearl, 1988].

For each $\sigma$ we generated and tested 100 samples divided into 10 different networks each simulated with 10 different input bit vectors [1] We also tried to run Elim-MPE on this set of problems, but the induced width $w^*$ was too large and Elim-MPE failed to solve any problems.

In Table 1 there are 5 horizontal blocks, each corresponding to a different value of channel noise $\sigma$. Each block reports a distribution over the 95% accuracy range. Within each block we have 3 rows, one for each of MB (mini-bucket), BBMB and BFMB. Columns 3 through 6 report the results on various i-bounds. Column 7 reports results for IBP.

Looking at the third block in Table 1 (corresponding to $\sigma = 0.32$), we see that MB with i=2 (column 3) solved 45% of the problems exactly ($opt \geq 0.95$), while taking 0.05 seconds on the average. On the same set of problems, using mini-bucket heuristics, BBMB solved 96% of the problems exactly while taking 0.94 seconds on the average, while BFMB solved all problems exactly with average time of 0.13 seconds only. When moving to the right to columns 4 through 6 in rows corresponding to $\sigma = 0.32$ and $opt \geq 0.95$ we see the gradual change caused by higher level of mini-bucket heuristic (higher values of i-bound). As expected, MB solves more problems, while using more time. Focusing on BFMB we see that it always solved all problems with any i-bound, and its total running time as a function of $i$ forms a U-shaped curve. At first (i=2) it is high (0.13), then as i-bound increases the total time decreases (when i=6 total time is 0.10), but then as i-bound increases further the total time starts to increase again.

The added amount of search on top of MB can be estimated by $t_{search} = t_{total} - t_{MB}$. For each value of $\sigma$, as $i$ increases the average search time $t_{search}$ decreases, and the overall accuracy of search increases (more problems fall within higher ranges of $opt$). However, as $i$ increases, the amount of MB preprocessing increases as well.

Each line in the table demonstrates the tradeoff between the amount of preprocessing performed by MB and the amount of subsequent search using the heuristic cost function generated by MB. We observe that the total time improves when $i$ increases until a threshold point and then worsens. When $i$ is smaller than this threshold, the heuristic cost function is weak and search takes longer. When $i$ is larger than this threshold, the extra preprocessing is not cost effective.

We observe that as problems become harder (i.e. $\sigma$ increases) both search algorithms achieve their best performance for larger $i$ when the mini-bucket heuristics is stronger. For example, in Table 1, when $\sigma$ is 0.22, the optimal performance is for $i = 2$. When $\sigma$ is 0.40, the optimal point is $i = 10$.

One crucial difference between BBMB and BFMB is that BBMB is an anytime algorithm - it always outputs an assignment, and as time increases, its solution gets better. BFMB on the other hand, is an all-or-nothing algorithm. It only outputs a solution when it finds an optimal solution. In our experiments, we always used a time bound. If BFMB did not finish within the time bound, it outputed the MB assignment. From Table 1 we see that when sufficient time is given (indicated by cases when both BBMB and BFMB solve all problems) the average running time of BFMB is never worse than BBMB and often better by a factor of 5-10.

---

[1] In the past ([Kask and Dechter, 1999a]) we have run a large number of random coding experiments with different variable orderings. The results we report in this paper (with min-degree ordering) are typical of all the experiments we have run.

| N=200 K=100 σ | opt | MB BBMB BFMB i=2 # / time | MB BBMB BFMB i=6 # / time | MB BBMB BFMB i=10 # / time | MB BBMB BFMB i=14 # / time | IBP |
|---|---|---|---|---|---|---|
| 0.22 | >0.95 | 79/0.09 98/0.94 100/0.12 | 84/0.12 98/0.65 100/0.16 | 90/0.73 99/0.95 100/0.77 | 95/8.00 100/8.04 100/8.03 | 100/0.16 |
| 0.28 | >0.95 | 41/0.09 84/3.72 100/0.80 | 45/0.12 88/4.17 100/0.56 | 59/0.72 96/2.50 100/0.89 | 71/7.95 99/8.64 100/8.03 | 100/0.13 |
| 0.32 | >0.95 | 18/0.09 63/10.2 94/6.19 | 21/0.12 68/9.49 96/4.12 | 31/0.73 87/6.33 99/2.49 | 46/8.06 92/10.7 100/8.75 | 99/0.16 |
| 0.40 | >0.95 | N/A N/A N/A | 0/- 28/90.5 39/66.7 | 5/0.73 39/51.8 67/50.1 | 6/7.77 58/42.7 85/41.1 | 77/0.15 |

Table 3: Random coding, N=200 K=100. 100 samples.

| N=200 K=100 σ | MB BBMB BFMB i=8 | MB BBMB BFMB i=10 | MB BBMB BFMB i=12 | MB BBMB BFMB i=14 | IBP |
|---|---|---|---|---|---|
| 0.22 | 0.004280 0.000680 0.000220 | 0.003780 0.000600 0.000220 | 0.002100 0.000340 0.000220 | 0.001200 0.000220 0.000220 | 0.000220 |
| 0.28 | 0.020260 0.007820 0.001020 | 0.020580 0.006040 0.001020 | 0.015320 0.002620 0.001020 | 0.009220 0.001360 0.001020 | 0.001040 |
| 0.32 | 0.042840 0.023780 0.006660 | 0.043740 0.021340 0.006060 | 0.038060 0.010980 0.003520 | 0.027600 0.007280 0.002820 | 0.002820 |
| 0.40 | N/A N/A N/A | 0.115200 0.083500 0.081200 | 0.105100 0.054900 0.050900 | 0.097300 0.047400 0.026400 | 0.011700 |

Table 4: Random coding BER, N=200 K=100. 100 samples.

In Table 2 we report the Bit Error Rate (BER) for the same problems and algorithms as in Table 1. BER is a standard measure used in the coding literature denoting the fraction of input bits that were decoded incorrectly. We observe that when the noise is very small (0.22, 0.28) BBMB and BFMB are equal to IBP since both BBMB/BFMB and IBP solve all problems exactly. However, when noise increases (0.51) BBMB and BFMB outperform IBP when the i-bound is sufficiently large. We ran 30 iterations of IBP on each problem and noticed that usually it converged to the final assignment after 5-10 iterations. Giving it more time would not improve its performance.

This phenomenon is more pronounced in Tables 3 and 4, where we present results with K=100 input bits. In this set of experiments we increased the time bound from 30 sec to 60 seconds (for small noise) or to 180 seconds (for large noise), while doubling the problem size. Again, we see a similar pattern of preprocessing-search tradeoff as with networks of K=50 bits. Also, we observe the superiority of BFMB over BBMB. Given the same i-bound, BFMB can solve more problems than BBMB and faster.

In Figures 7, 8 and 9 we provide an alternative view of the performance of BBMB(i) and BFMB(i). Let $F_{BBMB(i)}(t)$ ($F_{BFMB(i)}(t)$) be the fraction of the problems solved completely by BBMB(i) (BFMB(i)) by time $t$. Each graph in Figures 7, 8 and 9 plots $F_{BBMB(i)}(t)$ and $F_{BFMB(i)}(t)$ for some specific value of i.

Figures 7, 8 and 9 display a trade-off between preprocessing and search. Clearly, if $F_{BBMB(i)}(t) > F_{BBMB(j)}(t)$, then $F_{BBMB(i)}(t)$ completely dominates $F_{BBMB(j)}(t)$. For example, in Figure 9 BBMB(10)

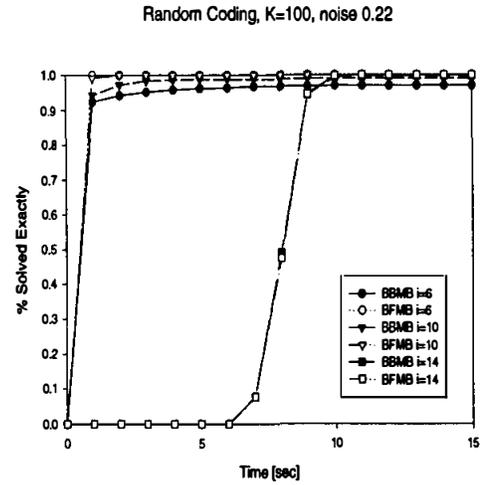

Figure 7: Random Coding. K=100, σ = 0.22.

completely dominates BBMB(6). When $F_{BBMB(i)}(t)$ and $F_{BBMB(j)}(t)$ intersect, they display a trade-off as a function of time. For example, if we have only few seconds, BBMB(6) is better than BBMB(14). However, when sufficient time is allowed, BBMB(14) is superior to BBMB(6).

Figures 7, 8 and 9 also show that $F_{BFMB(i)}(t)$ always dominates $F_{BBMB(i)}(t)$ for any value of i.

### 4.2 Random Noisy-OR Networks

Random Noisy-OR networks were randomly generated using parameters (N, K, C, P), where N is the number of variables, K is their domain size, C is the number of conditional probability matrices and P is the number of parents in each conditional probability matrix.

The structure of each test problem is created by randomly picking C variables out of N and for each, randomly selecting P parents from preceding vari-ables, relative to some ordering. Each probability table represents an OR-function with a given noise and leak probabilities : $P(X = 0|Y_1, ..., Y_P) = P_{leak} \times \Pi_{Y_i=1} P_{noise}$.

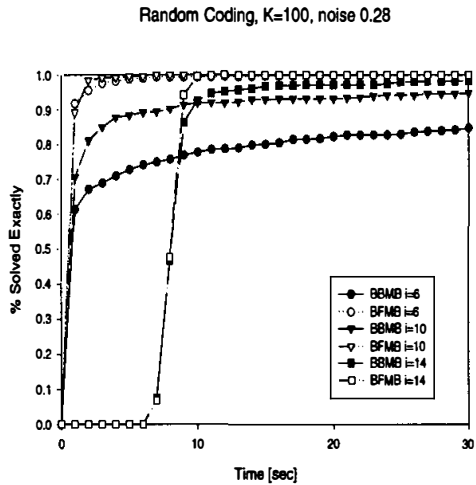

Figure 8: Random Coding. K=100, $\sigma = 0.28$.

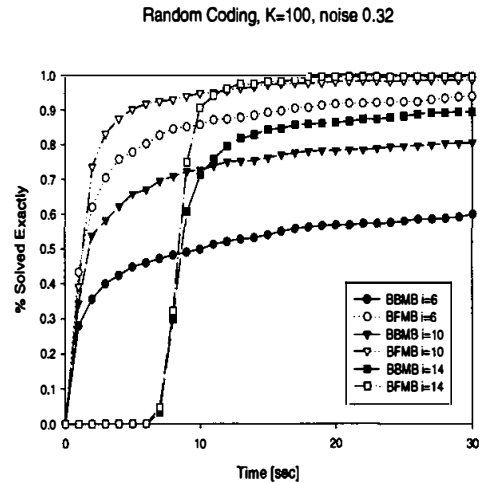

Figure 9: Random Coding. K=100, $\sigma = 0.32$.

| N C P | opt | MB BBMB BFMB i=2 # / time | MB BBMB BFMB i=6 # / time | MB BBMB BFMB i=10 # / time | MB BBMB BFMB i=14 # / time |
|---|---|---|---|---|---|
| 128 85 4 | >0.95 | 100/0.05 100/0.95 100/1.76 | 100/0.08 100/0.59 100/1.19 | 100/0.65 100/0.98 100/1.37 | 100/8.07 100/8.47 100/8.56 |
| 128 95 4 | >0.95 | 99/0.06 100/1.68 100/2.68 | 99/0.09 100/1.68 100/2.58 | 99/0.74 100/1.69 100/2.47 | 99/9.06 100/9.70 100/9.96 |
| 128 105 4 | >0.95 | 99/0.08 100/2.72 100/4.83 | 99/0.10 100/2.24 100/4.37 | 99/0.80 100/1.80 100/2.87 | 99/10.0 100/10.9 100/11.7 |

Table 5: Noisy-OR MPE. $P_{noise}=0.2$, $P_{leak}=0.01$. 10 evidence variables.

| CPCS360b 100 samples 10 evid. | MB BBMB BFMB i=4 | MB BBMB BFMB i=8 | MB BBMB BFMB i=12 | MB BBMB BFMB i=16 |
|---|---|---|---|---|
| >0.95 | 93[0.91] 100[0.93] 100[0.98] | 93[0.93] 100[0.94] 100[0.96] | 96[1.99] 100[2.00] 100[2.00] | 98[15.8] 100[15.8] 100[15.8] |
| CPCS422b 100 samples 10 evid. | MB BBMB BFMB i=4 | MB BBMB BFMB i=8 | MB BBMB BFMB i=12 | MB BBMB BFMB i=16 |
| >0.95 | 40[22.6] 96[24.8] 97[25.9] | 46[23.1] 98[24.5] 97[24.5] | 51[22.7] 100[22.9] 100[23.1] | 59[39.0] 100[39.0] 100[39.1] |

Table 6: CPCS networks. Time 30 and 45 resp.

Table 5 presents results of experiments with random Noisy-OR networks. Parameters N, K and P are fixed, while C, controlling network's sparseness, is changing.

Here we see a similar pattern of tradeoff between mini-bucket preprocessing and search. Mini-bucket algorithm can solve most of the problems exactly, but it takes a considerable amount of BBMB/BFMB search time to actually prove the optimality of the mini-bucket solution. We also see that here Branch and Bound is slightly faster than Best-First. This is because the lower bound used by BBMB is optimal from the beginning (MB solves the problem), the heuristic function is accurate ($f = f^*$) and there are many solutions. However, Best-First expands many more nodes before finding a solution because we use a random tie-breaking rule.

### 4.3 CPCS Networks

As another realistic domain, we used the CPCS networks derived from the Computer-Based Patient Care Simulation system, and based on INTERNIST-1 and Quick Medical Reference expert systems [Pradhan et al., 1994]. The nodes in CPCS networks correspond to diseases and findings. Representing it as a belief network requires some simplifying assumptions, 1) conditional independence of findings given diseases, 2) noisy-OR dependencies between diseases and findings, and 3) marginal independencies of diseases. For details see [Pradhan et al., 1994].

In Table 5 we have results of experiments with two binary CPCS networks, cpcs360b (N = 360, C = 335) and cpcs422b (N = 422, C = 348), with 100 instances in both cases. Each instance had 10 evidence nodes picked randomly.

Our results show a similar pattern of tradeoff between MB preprocessing and BBMB/BFMB search. Since cpcs360b network is solved quite effectively by the ap-proximation scheme MB, we get very good heuristics and therefore, the added search time is relatively small, serving primarily to prove the optimality of MB solution. On the other hand, on cpcs422b MB can solve less than half of the instances accurately when $i$ is small, and more as $i$ increases. BBMB/BFMB are roughly the same, both enhance MB's solution quality, significantly. They can solve all instances accurately for $i \geq 12$. For comparison, elim-mpe solved the cpcs360 network (with no evidence) in 115 sec while for cpcs422 it took 1697 sec. Processing the networks with evidence is a much more challenging task, however.

## 5 Discussion and Conclusion

Our experiments demonstrate the potential of mini-bucket heuristics in improving general search. The mini-bucket heuristic's accuracy can be controlled to yield an optimal tradeoff between preprocessing and search. We demonstrated this property in the context of both Branch-and-Bound [Kask and Dechter, 1999a] and Best-First search. Although the best threshold point cannot be predicted apriori a preliminary empirical analysis can be informative when given a class of problems that is not too heterogeneous.

The mini-bucket heuristics can facilitate Best-First search on relatively sizable problems, thus extending the boundaries of this search scheme which is computation optimal (relative to search algorithms having access to the same heuristic) for achieving exact solution. Indeed, we showed that Best-First usually outperforms Branch-and-Bound, sometimes by a factor of 5-10.

We showed that search can be competitive with the best known approximation algorithms for probabilistic decoding such as IBP when the networks are relatively small, in which case search solved the problems optimally. Obviously when problem sizes increase BBMB and BFMB require much more time. However, as much as IBP is efficient, its performance will not improve with time.

Finally, since mini-bucket elimination is applicable across many problem tasks such as probabilistic inference and decision making the scheme proposed here has potential of being widely applicable.